\documentclass[twoside,leqno,twocolumn]{article}

\usepackage[letterpaper]{geometry}
\usepackage{ltexpprt}
\usepackage{lmodern}
\usepackage{amssymb}
\usepackage{hyperref}
\usepackage{booktabs}
\usepackage[mathscr]{eucal}
\usepackage{graphicx}
\usepackage{mathtools} \mathtoolsset{showonlyrefs}
\usepackage{tabularx}
\usepackage{microtype}
\usepackage{bm}

\newtheorem{assumption}{Assumption}

\newcommand\Rb{\mathbb{R}}

\newcommand{\zerobf}[0]{\mathbf{0}}

\newcommand{\Abf}[0]{\mathbf{A}}

\newcommand{\Dbf}[0]{\mathbf{D}}

\newcommand{\Ibf}[0]{\mathbf{I}}

\newcommand{\Lbf}[0]{\mathbf{L}}
\newcommand{\Mbf}[0]{\mathbf{M}}

\newcommand{\Ubf}[0]{\mathbf{U}}

\newcommand{\Wbf}[0]{\mathbf{W}}
\newcommand{\Xbf}[0]{\mathbf{X}}
\newcommand{\Ybf}[0]{\mathbf{Y}}

\newcommand{\ubf}[0]{\mathbf{u}}
\newcommand{\vbf}[0]{\mathbf{v}}
\newcommand{\xbf}[0]{\mathbf{x}}

\newcommand{\thetabf}[0]{\boldsymbol{\theta}}
\newcommand{\Thetabf}[0]{\boldsymbol{\Theta}}

\newcommand{\Ae}{\boldsymbol{\mathscr{A}}}

\newcommand{\De}{\boldsymbol{\mathscr{D}}}

\newcommand{\Ie}{\boldsymbol{\mathscr{I}}}

\newcommand{\Le}{\boldsymbol{\mathscr{L}}}

\newcommand{\Qe}{\boldsymbol{\mathscr{Q}}}
\newcommand{\Se}{\boldsymbol{\mathscr{S}}}

\newcommand{\Ve}{\boldsymbol{\mathscr{V}}}
\newcommand{\We}{\boldsymbol{\mathscr{W}}}
\newcommand{\Xe}{\boldsymbol{\mathscr{X}}}
\newcommand{\Ye}{\boldsymbol{\mathscr{Y}}}

\DeclareMathOperator*{\triangleop}{\triangle}

\newcommand{\fold}{\operatorname{fold}}
\newcommand{\unfold}{\operatorname{unfold}}

\newcommand{\softmax}{\mathrm{softmax}}

\newcommand{\defeq}{\stackrel{\text{\tiny \textnormal{def}}}{=}}  

\newcolumntype{Y}{>{\centering\arraybackslash}X}
\newcommand{\tdag}{\textsuperscript{\textdagger}}
\newcommand{\tast}{\textsuperscript{\textasteriskcentered}}

\def\vh{{\bm{h}}}

\def\vm{{\bm{m}}}

\def\vz{{\bm{z}}}

\def\eve{{e}}

\def\mA{{\bm{A}}}

\DeclareMathAlphabet{\mathsfit}{\encodingdefault}{\sfdefault}{m}{sl}
\SetMathAlphabet{\mathsfit}{bold}{\encodingdefault}{\sfdefault}{bx}{n}

\def\gG{{\mathcal{G}}}

\newcommand{\ours}{\textsc{TM-GCN}}

\title{\Large Dynamic Graph Convolutional Networks Using the Tensor M-Product}

\author{
	Osman Asif Malik\thanks{University of Colorado Boulder,
	\texttt{osman.malik@colorado.edu}}
	\and
    Shashanka Ubaru\thanks{IBM Research,
    \texttt{shashanka.ubaru@ibm.com}}
    \and
    Lior Horesh\thanks{IBM Research,
    \texttt{lhoresh@us.ibm.com}}
    \and
    Misha E.~Kilmer\thanks{Tufts University,
    \texttt{misha.kilmer@tufts.edu}}
    \and
    Haim Avron\thanks{Tel Aviv University,
    \texttt{haimav@tauex.tau.ac.il}}
}
\date{}
\begin{document}

\maketitle

\fancyfoot[R]{\scriptsize{Copyright \textcopyright\ 2021 by SIAM\\
Unauthorized reproduction of this article is prohibited}}

\begin{abstract} \small\baselineskip=9pt
Many irregular domains such as social networks, financial transactions, neuron connections, and natural language constructs are represented using graph structures. In recent years, a variety of  graph neural networks (GNNs) have been successfully applied for representation learning and prediction on such graphs. In many of the real-world applications, the underlying graph changes over time, however, most of the existing GNNs are inadequate for handling such dynamic graphs.
In this paper we propose a novel technique for learning embeddings of dynamic graphs using a
tensor algebra framework. Our method extends the popular graph convolutional network (GCN) for learning representations of dynamic graphs using the recently proposed tensor M-product technique. Theoretical results presented establish a connection between the proposed tensor approach and spectral convolution of tensors. The proposed method \ours~ is consistent with the Message  Passing  Neural  Network  (MPNN) framework, accounting for both spatial and temporal message passing.  Numerical experiments on real-world datasets demonstrate the performance of the proposed method for edge classification and link prediction tasks on dynamic graphs. We also consider an application related to the COVID-19 pandemic, and show how our method can be used for early detection of infected individuals from  contact tracing data.
\end{abstract}

\section{Introduction}
Graphs are popular data structures used to effectively represent interactions and structural relationships between entities in structured data domains. Inspired by the success of deep neural networks for learning representations in the image and language domains, recently, application of neural networks for graph representation learning has attracted much interest. 
A number of  graph neural network (GNN) architectures have been 
explored in the contemporary literature for a variety of graph related tasks and applications~\cite{zhou2018,wu2019}. Methods based on graph convolution filters which extend convolutional neural networks (CNNs) to irregular graph domains are popular~\cite{bruna2013,defferrard2016,kipf2016}. Most of these GNN models operate on a given, static graph.

In many real-world applications, the underlying graph changes over time, and learning representations of such  dynamic graphs is essential. Examples include analyzing social networks~\cite{berger-wolf2006}, detecting fraud and crime in financial networks~\cite{pareja2019}, traffic control~\cite{zhao2019}, understanding neuronal activities in the brain~\cite{devicofallani2014}, and analyzing contact tracing data~\cite{ubaru2020dynamic}. In such dynamic settings, the temporal interdependence in the graph connections and features also play a substantial role.
However, efficient GNN methods that handle time varying graphs and that capture the temporal correlations are lacking.
 
By \emph{dynamic graph}, we refer to a sequence of graphs $\gG^{(t)}=(V, \Abf^{(t)}, \Xbf^{(t)})$, $t \in \{1, 2, \ldots, T\}$,  with a fixed set $V$ of $N$ nodes, adjacency matrices $\Abf^{(t)} \in \Rb^{N \times N}$, and graph feature matrices $\Xbf^{(t)} \in \Rb^{N \times F}$ where $\Xbf^{(t)}_{n:} \in \Rb^{F}$ is the feature vector consisting of $F$ features associated with node $n$ at time $t$. The graphs can be weighted, and directed or undirected. They can also have additional properties like (time varying) node and edge classes, which would be stored in a separate structure.
Suppose we only observe the first $T' < T$ graphs in the sequence. The goal of our method is to use these observations to predict some property of the remaining $T-T'$ graphs. In this paper, we consider edge classification, link prediction and node property prediction tasks.

\begin{figure}[t]
	\centering
	\includegraphics[width=1\linewidth]{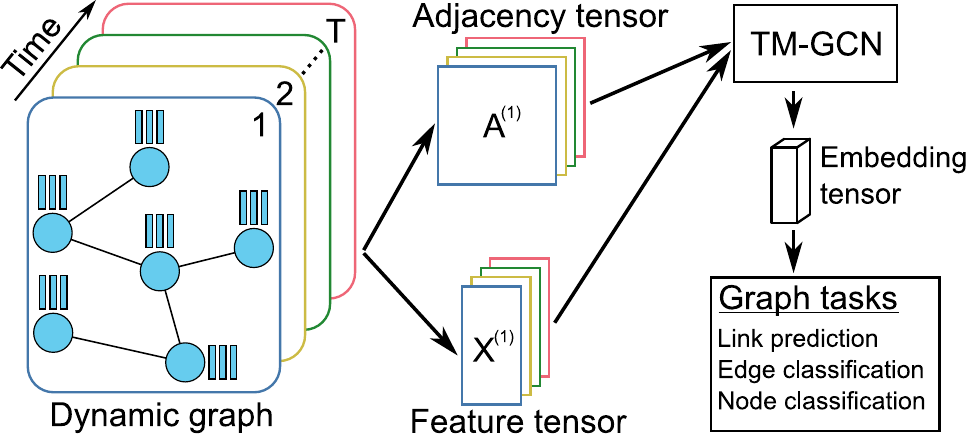}
	\caption{Our proposed TM-GCN approach.}
	\label{fig:TGCN}
\end{figure}

In recent years, tensor constructs have been explored to effectively process high-dimensional data,  in order to better leverage the multidimensional structure of such data~\cite{kolda2009}.
Tensor based approaches have been shown to perform well  in many applications.
Recently, a new tensor framework called the \emph{tensor M-product framework}~\cite{braman2010,kernfeld2015} was proposed that extends matrix based theory to high-dimensional architectures.

In this paper, we propose a novel tensor variant of the popular graph convolutional network (GCN) architecture~\cite{kipf2016}, which we call \ours. It captures correlation over time by leveraging the tensor M-product framework. The flexibility and matrix mimeticability of the framework, help us adapt the GCN architecture to tensor space.
Figure~\ref{fig:TGCN} illustrates our method at a high level: First, the time varying adjacency matrices $\Abf^{(t)}$ and  feature matrices $\Xbf^{(t)}$ of the dynamic graph are aggregated into an adjacency 
tensor and a feature tensor, respectively. 
These tensors are then fed into our \ours, which computes an embedding that can be used for a variety of tasks, such as link prediction, and edge and node classification. 
GCN architectures are motivated by graph convolution filtering, i.e., applying filters/functions to the graph Laplacian~\cite{bruna2013}, and we establish a similar connection between \ours~  and spectral filtering of tensors. Such results suggest possible extensions of other convolution based GNNs such as~\cite{bruna2013,defferrard2016} for dynamic graphs using the tensor framework. 
The Message  Passing  Neural  Network  (MPNN) framework has been used to describe spatial convolution GNNs~\cite{gilmer2017neural}. We show that \ours~ is consistent with the MPNN framework, and accounts for spatial and temporal message passing. 
Experimental results on real datasets illustrate the performance of our method for the edge classification and link prediction tasks on dynamic graphs. We also demonstrate how \ours~ can be used in an important application related to the COVID-19 pandemic. We show how GNNs can be used for early identification of individuals who are infected (potentially before they display symptoms) from contact tracing data and a dynamic graph based SEIR model~\cite{ubaru2020dynamic}.

\section{Related Work}
\paragraph{Unsupervised Embedding:} Unsupervised graph embedding techniques have been popular  for  link prediction on static graphs~\cite{cai2018comprehensive}. A number of dynamic graph embedding 
methods have been proposed recently, which extend the static ones. 
DANE~\cite{li2017attributed} adapted the popular dimensionality reduction approaches such as Eigenmaps to time varying graphs by efficiently updating the eigenvectors from the prior ones.
The popular random walk based methods have also been extended to  obey the temporal order in recent works~\cite{nguyen2018continuous}.

Numerous deep neural network based unsupervised learning methods have been 
developed for dynamic graph embedding. Examples include DynGEM~\cite{goyal2018dyngem},
Know-Evolve~\cite{trivedi2017know}, DyRep~\cite{trivedi2019dyrep},  Dynamic-Triad~\cite{zhou2018dynamic}, 
and others. In most of these methods, a temporal smoothness regularization is used to obtain stable embedding across consecutive time-steps.

\paragraph{Supervised Learning:}
The idea of using graph  convolution  based  on the  spectral  graph  theory for GNNs was first introduced by \cite{bruna2013}. 
\cite{defferrard2016} then proposed \emph{Chebnet}, where the spectral filter was approximated by Chebyshev polynomials in order to make it faster and localized. 
\cite{kipf2016} presented the simplified GCN, a degree-one polynomial approximation of Chebnet, in order to speed up computation further and improve the performance.
There are many other works that deal with GNNs when the graph and features are fixed/static; see the review papers~\cite{zhou2018} and \cite{wu2019} and references therein.

Recently, Li et.~al~\cite{li2017} develop a diffusion convolutional RNN for traffic forecasting, where road networks are modeled assuming both the nodes and edges remain fixed over time, unlike in our setting.
Seo et.~al~\cite{seo2018} devise the Graph Convolutional Recurrent Network for graphs with time varying features, while the edges are fixed over time. EdgeConv was proposed in~\cite{wang2018a}, which is a neural network (NN) approach that applies convolution operations on static graphs in a dynamic fashion. 
\cite{zhao2019} develop a temporal GCN method called T-GCN, which they apply for traffic prediction. Here too, the graph remains fixed over time, and only the features vary. 
\cite{zhang2018a} propose a method which they refer to as a tensor graph CNN. Here, the standard GCN~\cite{kipf2016} based on matrix algebra is considered, and a ``cross graph convolution'' layer is introduced to handle the time varying aspect of the dynamic graph. In particular, the cross graph convolution layer involves computing a parameterized Kronecker sum  of the current adjacency matrix with the previously processed adjacency matrix, followed by a GCN layer. Recently, \cite{liu2020tensor} described a tensor version of GCN for text classification, where the text semantics are represented as a three-dimensional graph tensor. This work neither considers time varying graphs, nor the tensor M-product framework.

The set of methods most relevant to our setting of learning embeddings of dynamic graphs use combinations of GNNs and recurrent architectures (RNN), to capture the graph structure and handle time dynamics, respectively.
The approach in \cite{manessi2019} uses Long Short-Term Memory (LSTM), a recurrent network, in order to handle time variations along with GNNs. They design architectures for semi-supervised node classification and for supervised graph classification.
\cite{pareja2019} presented a variant of GCN called \emph{EvolveGCN}, where Gated Recurrent Units (GRUs) and LSTMs are coupled with a GCN to handle dynamic graphs. This paper is currently the state-of-the-art.  \cite{sankar2018dynamic} proposed the use of a temporal self-attention layer for dynamic graph representation learning. However, all these approaches are based on a heuristic RNN/GRU mechanism to evolve weights, and the models are not time aware (time is not an explicit entity).
\cite{newman2018} present a tensor NN which utilizes the tensor M-product  framework. Their approach is applicable  to image and other high-dimensional data that lie on regular grids.

\section{Tensor M-Product Framework}

Here, we cover the necessary preliminaries on tensors and the M-product framework. For a more general introduction to tensors, we refer the reader to the review paper  \cite{kolda2009}. In the present paper, a \emph{tensor} is a three-dimensional array of real numbers denoted by boldface Euler script letters, e.g.\ $\Xe \in \Rb^{I \times J \times T}$. Matrices are denoted by bold uppercase letters, e.g.\ $\Xbf$; vectors are denoted by bold lowercase letter, e.g.\ $\xbf$; and scalars are denoted by lowercase letters, e.g.\ $x$. An element at position $(i,j,t)$ in a tensor is denoted by subscripts, e.g.\ $\Xe_{ijt}$, with similar notation for elements of matrices and vectors. A colon will denote all elements along that dimension; $\Xbf_{i:}$ denotes the $i$th row of the matrix $\Xbf$, and $\Xe_{::k}$ denotes the $k$th frontal slice of $\Xe$. The vectors $\Xe_{ij:}$ are called the \emph{tubes} of $\Xe$.

The framework we consider relies on a new definition of the product of two tensors, called the M-product \cite{braman2010, kilmer2013, kernfeld2015}. A distinguishing feature of this framework is that the M-product of two three-dimensional tensors is also three-dimensional, which is not the case for e.g.\ tensor contractions~\cite{kolda2009}. It allows one to elegantly generalize many classical numerical methods from linear algebra. 
The framework, originally developed for three-dimensional tensors, has been extended to handle tensors of dimension greater than three~\cite{kilmer2013}. 
The following definitions~\ref{def:M-transform}--\ref{def:M-product} describe the M-product.
\begin{Definition}[M-transform] \label{def:M-transform}
    Let $\Mbf \in \Rb^{T \times T}$ be a mixing matrix.
    The \emph{M-transform} of a tensor $\Xe \in \Rb^{I \times J \times T}$ is denoted by $\Xe \times_3 \Mbf \in \Rb^{I \times J \times T}$ and defined elementwise as 
    \begin{equation} \label{eq:M-transform}
        (\Xe \times_3 \Mbf)_{ijt} \defeq \sum_{k = 1}^T \Mbf_{tk} \Xe_{ijk}.
    \end{equation}
    We say that $\Xe \times_3 \Mbf$ is in the \emph{transformed space}. Note that if $\Mbf$ is invertible, then $(\Xe \times_3 \Mbf) \times_3 \Mbf^{-1} = \Xe$. Consequently, $\Xe \times_3 \Mbf^{-1}$ is the \emph{inverse M-transform} of $\Xe$. The definition in \eqref{eq:M-transform} may also be written in matrix form as $\Xe \times_3 \Mbf \defeq \fold(\Mbf \unfold(\Xe))$, where the unfold operation takes the tubes of $\Xe$ and stack them as columns into a $T \times IJ$ matrix, and $\fold(\unfold(\Xe)) = \Xe$.
    
\end{Definition}

\begin{Definition}[Facewise product] \label{def:facewise-product}
    Let $\Xe \in \Rb^{I \times J \times T}$ and $\Ye \in \Rb^{J \times K \times T}$ be two tensors. The \emph{facewise product}, denoted by $\Xe \triangleop \Ye \in \Rb^{I \times K \times T}$, is defined facewise as $(\Xe \triangleop \Ye)_{::t} \defeq \Xe_{::t} \Ye_{::t}$.
\end{Definition}
\begin{Definition}[M-product] \label{def:M-product}
Let $\Xe \in \Rb^{I \times J \times T}$ and $\Ye \in \Rb^{J \times K \times T}$ be two tensors, and let $\Mbf \in \Rb^{T \times T}$ be an invertible matrix. The \emph{M-product}, denoted by $\Xe \star \Ye \in \Rb^{I \times K \times T}$, is defined as
\begin{equation}
    \Xe \star \Ye \defeq ((\Xe \times_3 \Mbf) \triangleop (\Ye \times_3 \Mbf)) \times_3 \Mbf^{-1}.
\end{equation}
\end{Definition}
In the original formulation of the M-product, $\Mbf$ was chosen to be the Discrete Fourier Transform (DFT) matrix, which allows efficient computation using the Fast Fourier Transform (FFT) \cite{braman2010, kilmer2013}. The framework was later extended for arbitrary invertible $\Mbf$ (e.g.\ discrete cosine and wavelet transforms) \cite{kernfeld2015}. Additional details are in the supplement.

\section{Tensor Dynamic Graph Embedding} \label{sec:our-method}

Our approach is inspired by the first order GCN by \cite{kipf2016} for static graphs, owed to  its simplicity and effectiveness. For a graph with adjacency matrix $\Abf$ and feature matrix $\Xbf$, a  GCN layer takes the form $\Ybf = \sigma(\tilde{\Abf} \Xbf \Wbf)$, where 
\begin{equation} \label{eq:A-tilde}
    \tilde{\Abf} \defeq \tilde{\Dbf}^{-1/2} (\Abf + \Ibf) \tilde{\Dbf}^{-1/2},
\end{equation}
$\tilde{\Dbf}$ is diagonal with $\tilde{\Dbf}_{ii} = 1 + \sum_j \Abf_{ij}$, $\Ibf$ is the matrix identity, $\Wbf$ is a matrix to be learned when training the NN, and $\sigma$ is an activation function, e.g., ReLU.
Our approach translates this to a tensor model by utilizing the M-product framework. We first introduce a tensor activation function $\hat{\sigma}$ which operates in the transformed space.
\begin{Definition}
    Let $\Ae \in \Rb^{I \times J \times T}$ be a tensor and $\sigma$ an elementwise activation function. We define the activation function $\hat{\sigma}$ as $\hat{\sigma}(\Ae) \defeq \sigma(\Ae \times_3 \Mbf) \times_3 \Mbf^{-1}$.
\end{Definition}
We can now define our proposed dynamic graph embedding. Let $\Ae \in \Rb^{N \times N \times T}$ be a tensor with frontal slices $\Ae_{::t} = \tilde{\Abf}^{(t)}$, where $\tilde{\Abf}^{(t)}$ is the normalization of $\Abf^{(t)}$. Moreover, let $\Xe \in \Rb^{N \times F \times T}$ be a tensor with frontal slices $\Xe_{::t} = \Xbf^{(t)}$. Finally, let $\We \in \Rb^{F \times F' \times T}$ be a weight tensor. We define our dynamic graph embedding as $\Ye = \Ae \star \Xe \star \We \in \Rb^{N \times F' \times T}$. This computation can also be repeated in multiple layers. For example, a 2-layer formulation would be of the form \begin{equation}\Ye = \Ae \star \hat{\sigma}(\Ae \star \Xe \star \We^{(0)}) \star \We^{(1)}.
\end{equation}

One important consideration is how to choose the matrix $\Mbf$ which defines the M-product. 
For time-varying graphs, we choose $\Mbf$ to be lower triangular and banded so that each frontal slice $(\Ae \times_3 \Mbf)_{::t}$ is a linear combination of the adjacency matrices $\Ae_{::\max(1,t-b+1)}, \ldots, \Ae_{::t}$, where we refer to $b$ as the ``bandwidth'' of $\Mbf$. 
This choice ensures that each frontal slice $(\Ae \times_3 \Mbf)_{::t}$ only contains information from current and past graphs that are close temporally.
We consider two variants of the lower banded triangular $\Mbf$ matrix in the experiments; see the supplement for details.
Another possibility is to treat $\Mbf$ as a parameter matrix to be learned from the data. 

In order to avoid over-parameterization and improve the performance, we choose the  weight tensor $\We$ (at each layer), such that each of the frontal slices of $\We$ in the transformed domain remains the same, i.e., $(\We \times_3 \Mbf)_{::t} = (\We \times_3 \Mbf)_{::t'}\: \forall t, t'$. In other words, the parameters in each layer are shared and learned over all the training instances. This reduces the number of parameters to be learned significantly.  

An embedding $\Ye \in \Rb^{N \times F' \times T}$ can now be used for various prediction tasks, like link prediction, and edge and node classification. In Section~\ref{sec:experiments}, we apply our method for edge classification and link prediction by using a model similar to that used by \cite{pareja2019}: Given an edge between nodes $m$ and $n$ at time $t$, the predictive model is 
\begin{equation} \label{eq:prediction-model}
    p(m,n,t) \defeq \softmax(\Ubf [(\Ye \times_3 \Mbf)_{m:t}, (\Ye \times_3 \Mbf)_{n:t}]^\top),
\end{equation}
where $(\Ye \times_3 \Mbf)_{m:t} \in \Rb^{F'}$ and $(\Ye \times_3 \Mbf)_{n:t} \in \Rb^{F'}$ are row vectors, $\Ubf \in \Rb^{C \times 2F'}$ is a weight matrix, and $C$ the number of classes. Note that the embedding $\Ye$ is first M-transformed before the matrix $\Ubf$ is applied to the appropriate feature vectors. This, combined with the fact that the tensor activation functions are applied elementwise in the transformed domain, allow us to avoid ever needing to apply the inverse M-transform. This approach reduces the computational cost, and has been found to improve performance in the edge classification task.

\subsection{Theoretical Motivation for \ours} \label{sec:theoretical-analysis}

Here, we present the results that establish the connection between the
proposed TM-GCN and spectral convolution of tensors, in particular spectral filtering and approximation on dynamic graphs. This is analogous to the graph convolution based on spectral graph theory in the GNNs by~\cite{bruna2013}, \cite{defferrard2016}, and \cite{kipf2016}.  All proofs and additional details are provided in Section~\ref{sec:proofs} of the supplement.

Let $\Le \in \Rb^{N \times N \times T}$ be a form of tensor Laplacian defined as $\Le \defeq \Ie - \Ae$.
Throughout the remainder of this subsection, we will assume that the adjacency matrices $\Abf^{(t)}$ are symmetric.
\begin{proposition} \label{prop:existence-eigendecomp}
    The tensor $\Le$ has an eigendecomposition $\Le = \Qe \star \De \star \Qe^{\top}$.
\end{proposition}

\begin{Definition}[Filtering]
    Given a signal $\Xe \in \Rb^{N \times 1 \times T}$ and a function $g : \Rb^{1 \times 1 \times T} \rightarrow \Rb^{1 \times 1 \times T}$, we define the \emph{tensor spectral graph filtering} of $\Xe$ with respect to $g$ as
    \begin{equation} \label{eq:X-filt}
        \Xe_\textup{filt} \defeq \Qe \star g(\De) \star \Qe^\top \star \Xe,
    \end{equation}
    where
    \begin{equation}
        g(\De)_{mn:} \defeq
        \begin{cases}
            g(\De_{mn:}) & \text{if } m=n, \\
            \zerobf & \text{if } m \neq n.
        \end{cases}
    \end{equation}
\end{Definition}
In order to avoid the computation of an eigendecomposition, \cite{defferrard2016} uses a polynomial to approximate the filter function. We take a similar approach, and approximate $g(\De)$ with an M-product polynomial. For this approximation, we impose additional structure on $g$.
\begin{assumption} \label{as:g}
    Assume that $g : \Rb^{1 \times 1 \times T} \rightarrow \Rb^{1 \times 1 \times T}$ is defined as
    \begin{equation}
        g(\Ve) \defeq f(\Ve \times_3 \Mbf) \times_3 \Mbf^{-1},
    \end{equation}
    where $f$ is defined elementwise as $f(\Ve \times_3 \Mbf)_{11t} \defeq f^{(t)}((\Ve \times_3 \Mbf)_{11t})$ with each $f^{(t)}:\Rb \rightarrow \Rb$ continuous.
\end{assumption}
\begin{proposition} \label{prop:tensor-polynomial-approx}
    Suppose $g$ satisfies Assumption~\ref{as:g}. For any $\varepsilon > 0$, there exists an integer $K$ and a set $\{\thetabf^{(k)}\}_{k=1}^K \subset \Rb^{1 \times 1 \times T}$ such that
    \begin{equation} \label{eq:tensor-polynomial-approx}
        \Big\| g(\De) - \sum_{k=0}^K \De^{\star k} \star \thetabf^{(k)} \Big\| < \varepsilon,
    \end{equation}
    where $\| \cdot \|$ is the tensor Frobenius norm, and where $\De^{\star k } \defeq \De \star \cdots \star \De$ is the M-product of $k$ instances of $\De$, with the convention that $\De^{\star 0} = \Ie$.
\end{proposition}
As in the work of \cite{defferrard2016}, a tensor polynomial approximation allows us to approximate $\Xe_\text{filt}$ in \eqref{eq:X-filt} without computing the eigendecomposition of $\Le$:
\begin{equation} \label{eq:X-filt-approx}
\begin{aligned}
    \Xe_\text{filt} 
    &= \Qe \star g(\De) \star \Qe^\top \star \Xe \\
    &\approx \Qe \star \Big( \sum_{k=0}^K \De^{\star k} \star \thetabf^{(k)} \Big) \star \Qe^\top \star \Xe \\ 
    &= \Big( \sum_{k=0}^K \Le^{\star k} \star \thetabf^{(k)} \Big) \star \Xe.
\end{aligned}
\end{equation}
All that is necessary is to compute tensor powers of $\Le$.
We can also define tensor polynomial analogs of the Chebyshev polynomials and do the approximation in \eqref{eq:X-filt-approx} in terms of those instead of the tensor monomials $\De^{\star k}$.  We note that if a degree-one approximation is used, the computation in \eqref{eq:X-filt-approx} becomes
\begin{equation}
\begin{aligned}
    \Xe_\text{filt} 
    &\approx (\Ie \star \thetabf^{(0)} + \Le \star \thetabf^{(1)}) \star \Xe \\
    &= (\Ie \star \thetabf^{(0)} + (\Ie - \Ae) \star \thetabf^{(1)}) \star \Xe.
\end{aligned}
\end{equation}
Setting $\thetabf \defeq \thetabf^{(0)} = -\thetabf^{(1)}$, which is analogous to the parameter choice made in the degree-one approximation in \cite{kipf2016}, we get
\begin{equation} \label{eq:single-feature-filter}
    \Xe_\text{filt} \approx \Ae \star \Xe \star \thetabf.
\end{equation}
If we let $\Xe$ contain $F$ signals, i.e., $\Xe \in \Rb^{N \times F \times T}$, and apply $F'$ filters, \eqref{eq:single-feature-filter} becomes
\begin{equation}
    \Xe_\text{filt} \approx \Ae \star \Xe \star \Thetabf \in \Rb^{N \times F' \times T},
\end{equation}
where $\Thetabf \in \Rb^{F \times F' \times T}$. This is precisely our embedding model, with $\Thetabf$ replaced by a learnable parameter tensor $\We$. These results show: (a) the connection between \ours~ and  spectral convolution of tensors, analogous to the GCN, and (b) that we can indeed develop higher order convolutional GNNs like~\cite{bruna2013,defferrard2016} for dynamic graphs using our framework.

\subsection{Message Passing Framework}
The Message  Passing  Neural  Network  (MPNN) framework is popularly used to describe spatial convolution GNNs~\cite{gilmer2017neural}.  
The graph convolution operation is considered to be a message passing process, with information being  passed  from  one  node  to  another  along the edges. The message passing phase of MPNN constitutes updating the hidden state $\vh_{v,\ell}$ at node $v$ in the $\ell$th layer with message $\vm_{v,\ell+1}$ as 
\begin{align}
    \vm_{v,\ell+1} &= \sum_{w\in N(v)}\Phi_{\ell}(\vh_{v,\ell},\vh_{w,\ell},\eve_{vw}),\\
    \vh_{v,\ell+1} & = \Psi_{\ell}(\vh_{v,\ell},\vm_{v,\ell+1}),
 \end{align}
where $N(v)$ is the neighbors of $v$ in the graph, $\eve_{vw}$ is the edge between nodes $v$ and $w$, $\Phi_{\ell}$ is a message function, and $\Psi_{\ell}$ is an update function. A number of GNN models  can be defined using this standard MPNN framework for static graphs. For the standard GCN model~\cite{kipf2016}, we have
$\Phi_{\ell}(\vh_{v,\ell},\vh_{w,\ell}) = A_{v,w} \vh_{w,\ell}$, where $A_{v,w} $ is the entry of adjacency matrix $\mA$, and $\Psi_{\ell}(\vh_{v,\ell},\vm_{v,\ell+1}) = \sigma(\vm_{v,\ell+1})$ where $\sigma$ is a pointwise non-linear function, e.g., ReLU. 

In this paper, we consider dynamic graphs and the designed GNN has to do spatial and temporal message passing. That is, for a graph $\gG^{(t)}$ at time $t$, the MPNN should be modeled such that the information/message is passed between neighboring nodes, as well as the corresponding nodes in the graphs $\{\gG^{(t-1)},\gG^{(t-2)}, \ldots, \gG^{(1)}\}$. Recently, a spatio-temporal message passing framework was defined for video processing in computer vision~\cite{mavroudi2019neural}. However, their framework does not account for time direction, and the graph is not considered to be evolving.
We define the message passing framework for a dynamic graph as follows:
The message passing phase will constitute updating the hidden state $\vh^{(t)}_{v,\ell}$ at node $v$ of graph $\gG^{(t)}$ in the $\ell$th layer with message $\vm^{(t)}_{v,\ell+1}$ as 
\begin{align}
    \vm^{(t)}_{v,\ell+1} &= \sum_{w\in N(v)}\sum_{\tau=1}^{t}\Phi^{(\tau)}_{\ell}
    \left(\sum_{\tau=1}^{t}\Gamma^{(\tau)}_{\ell}(\vh^{(\tau)}_{v,\ell}),
    \vh^{(\tau)}_{w,\ell},e^{\tau}_{vw}\right),\\
    \vh^{(t)}_{v,i+1} & = \Psi_{\ell}(\vh^{(t)}_{v,\ell},\vm^{(t)}_{v,\ell+1}).
 \end{align}
 Here, the function $\Gamma^{(\tau)}_{\ell}$ accounts for the message passing between hidden states over different time $\tau\leq t$, and function $\phi^{(\tau)}_{\ell}$ accounts for message passing between neighbors $N(v)$ over time $\tau\leq t$. 
 The model accounts for extensive spatio-temporal message passing.

For the proposed TM-GCN model, we have a function $\Gamma^{(\tau)}_{\ell}(\vh^{(\tau)}_{v,\ell}) = M_{t,\tau}\vh^{(\tau)}_{v,\ell}$, where $M_{t,\tau}$ is the $(t,\tau)$ entry of the mixing matrix $\Mbf$. The message function is $\Phi^{(\tau)}_{\ell}(\vz^{(t)}_{v,\ell},\vh^{(\tau)}_{w,\ell})
= M_{t,\tau} A_{v,w,\tau}$, where $\vz^{(t)}_{v,\ell} = \sum_{\tau=1}^{t}\Gamma^{(\tau)}(\vh^{(\tau)}_{v,\ell})$, and 
$A_{v,w,\tau}$ is the entry of the adjacency tensor $\Ae$.
The update function is $\Psi_{\ell}(\vh^{(t)}_{v,\ell},\vm^{(t)}_{v,\ell+1}) = \sigma(\vm^{(t)}_{v,\ell+1})$ with an elementwise non-linear function $\sigma$.

Note that the above message passing model does not include the inverse transform $\Mbf^{-1}$ as in the definition of the M-products. This is because, the M-transform is responsible for the temporal message passing and undoing it is not necessary. In our experiments too, we found that transforming back (applying  the inverse transform $\Mbf^{-1}$) did not yield improved results as suggested by the above MPNN model. This does not affect any of the theory presented in the previous section since the spectral filtering is performed in the transformed domain (see the supplement for details).
\begin{table}
\centering
    \caption{Dataset statistics.  \label{tab:data-stats}}
    \resizebox{0.5\textwidth}{!}{
    \begin{tabular}{lrrrrrrrr}
    \toprule
    & & & & Window & & \multicolumn{3}{c}{Partitioning}  \\
    \cmidrule(l){7-9}
    Dataset         & Nodes     & Edges     & $T$ &(days)  & $C$   & $S_\text{train}$ & $S_\text{val}$ & $S_\text{test}$ \\ 
    \midrule
    SBM     & 1,000    & 1,601,999    & 50          & --      & --     & 35 & 5 & 10 \\
    BitcoinOTC     & 6,005     & 35,569    & 135           & 14       & 2     & 95 & 20 & 20 \\
    BitcoinAlpha   & 7,604     & 24,173    & 135           & 14       & 2     & 95 & 20 & 20 \\
    Reddit          & 3,818     & 163,008   & 86            & 14        & 2     & 66 & 10 & 10 \\
    Chess           & 7,301     & 64,958    & 100           & 31        & 3     & 80 & 10 & 10 \\
    \bottomrule
    \end{tabular}
    }
\end{table}
\begin{figure}
	\centering  
	\includegraphics[width=.5\textwidth]{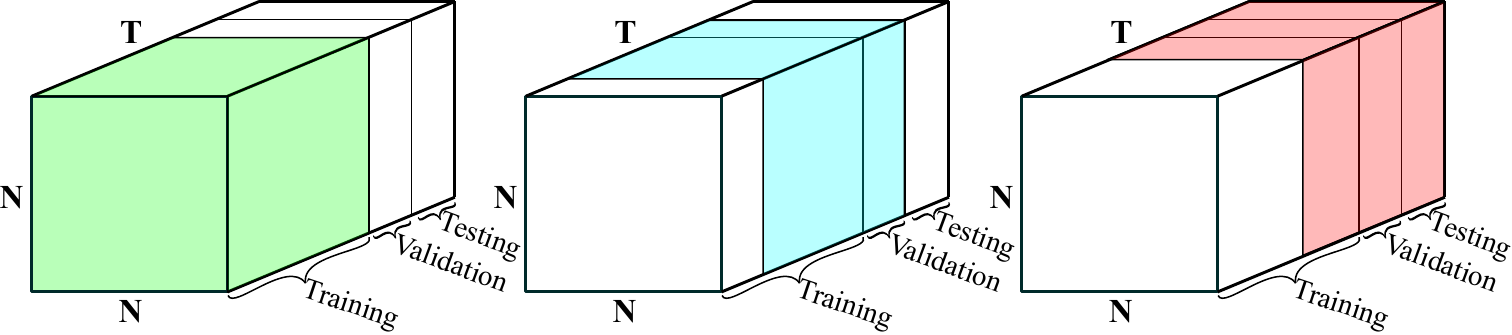}
	\caption{Partitioning of $\Ae$ into training, validation and testing data.}
	\label{fig:data-split}
\end{figure}

\section{Numerical Experiments} \label{sec:experiments}

\begin{table*}[tb!]
\centering
    \caption{Results for edge classification. Performance measures are F1 score\tdag or accuracy\tast. A higher value is better.\label{tab:results-asymmetric}}
    \begin{tabularx}{.9\textwidth}{lYYYY}
    \toprule
    & \multicolumn{4}{c}{Dataset}\\
    \cmidrule(l){2-5}
    Method                  & Bitcoin OTC\tdag  & Bitcoin Alpha\tdag& Reddit\tdag       & Chess\tast                \\
    \midrule
    WD-GCN                          & 0.3562            & 0.2533            & \textbf{0.2337}   & 0.4311            \\
    EvolveGCN                       & 0.3483            & 0.2273            & 0.2012            & 0.4351            \\
    GCN                             & 0.3402            & 0.2381            & 0.1968            & 0.4342            \\
    \ours~- M1    & 0.3660            & \textbf{0.3243}   & 0.2057            & \textbf{0.4708}   \\
    \ours~- M2    & \textbf{0.4361}   & 0.2466            & 0.1833            & 0.4513            \\
    \bottomrule
    \end{tabularx}
    
    \caption{Results for link prediction. Performance measure is MAP. A higher value is better. \label{tab:link-prediction}}
    \begin{tabularx}{.9\textwidth}{lYYYYY}
    \toprule
    & \multicolumn{5}{c}{Dataset}\\
    \cmidrule(l){2-6}
    Method                 &SBM  & Bitcoin OTC       & Bitcoin Alpha     & Reddit            & Chess             \\
    \midrule
    WD-GCN            &   0.9436   & 0.8071            & 0.8795            & \textbf{0.3896}   & 0.1279            \\
    EvolveGCN          &0.7620     & 0.6985            & 0.7722            & 0.2866            & 0.0915            \\
    GCN               &0.9201      & 0.6847            & 0.7655            & 0.3099            & 0.0899            \\
    \ours~- M1        &0.9684    & 0.8026            & 0.9318            & 0.2270            & \textbf{0.1882}   \\ 
   \ours~- M2      &\textbf{0.9799}      & \textbf{0.8458}   & \textbf{0.9631}   & 0.1405            & 0.1514            \\
    \bottomrule
    \end{tabularx}
\end{table*}
We first present results for edge classification and link prediction. 
We then show how we can use GNNs for predicting the state of individuals from COVID-19 contact tracing graphs. 

\subsection{Datasets and Preprocessing} We consider  five datasets  (links to the datasets are in the supplement): The Bitcoin Alpha and OTC transaction datasets \cite{pareja2019}, the Reddit body hyperlink dataset \cite{kumar2018}, a chess results dataset \cite{kunegis2013}, and SBM is the structure block matrix by~\cite{goyal2018dyngem}. The bitcoin datasets consist of transaction histories for users on two different platforms. Each node is a user, and each directed edge indicates a transaction and is labeled with an integer between $-10$ and $10$ which indicates the senders trust for the receiver. We convert these labels to two classes: positive (trustworthy) and negative (untrustworthy). The Reddit dataset is built from hyperlinks from one subreddit to another. Each node represents a subreddit, and each directed edge is an interaction which is labeled with $-1$ for a hostile interaction or $+1$ for a friendly interaction. We only consider those subreddits which have a total of 20 interactions or more. In the chess dataset, each node is a player, and each directed edge represents a match with the source node being the white player and the target node being the black player. Each edge is labeled $-1$ for a black victory, $0$ for a draw, and $+1$ for a white victory. Table~\ref{tab:data-stats} summarizes the statistics for the different datasets, where $T$ is total \# graphs and $C$ is the \# classes. The SBM dataset has no labels and hence we use it only for link prediction.

The data is temporally partitioned into $T$ graphs, with each graph containing data from a particular time window. Both $T$ and the time window length can vary between datasets. For each node-time pair $(n, t)$ in these graphs, we compute the number of outgoing and incoming edges and use these two numbers as features. The adjacency tensor $\Ae$ is then constructed as described in Section~\ref{sec:our-method}. The $T$ frontal slices of $\Ae$ are divided into $S_\text{train}$ training slices, $S_\text{val}$ validation slices, and $S_\text{test}$ testing slices, which come sequentially after each other; see Figure~\ref{fig:data-split} and Table~\ref{tab:data-stats}.

Since the adjacency matrices corresponding to graphs are very sparse for these datasets, we apply the same technique as \cite{pareja2019} and add the entries of each frontal slice $\Ae_{::t}$ to the following $l-1$ frontal slices $\Ae_{::t}, \ldots, \Ae_{::(t+l-1)}$, where we refer to $l$ as the ``edge life.''  Note that this only affects $\Ae$, and that the added edges are not treated as real edges in the classification and prediction problems.

The bitcoin and Reddit datasets are heavily skewed, with about 90\% of edges labeled positively, and the remaining labeled negatively. Since the negative instances are more interesting to identify (e.g.\ to prevent financial fraud or online hostility), we use the F1 score to evaluate the edge classification experiments on these datasets, treating the negative edges as the ones we want to identify. The classes are more well-balanced in the chess dataset, so we use accuracy to evaluate those edge classification experiments. 

\subsection{Graph Tasks}
For the link prediction experiments, we follow \cite{pareja2019} and use negative sampling to construct non-existing edges, and use mean average precision (MAP) as a performance measure. The negative sampling is done so that 5\% of edges are existing edges for each time slice, and all other edges are non-existing. Precise definitions of the different performance measures we use are given in Section~\ref{sec:hyperparameter-tuning} of the supplement.

For edge classification, we use an embedding $\Ye_\text{train} = \Ae_{::(1:S_\text{train})} \star \Xe_{::(1:S_\text{train})} \star \We$ for training. When computing the embeddings for the validation and testing data, we still need $S_\text{train}$ frontal slices of $\Ae$, which we get by using a sliding window of slices. This is illustrated in Figure~\ref{fig:data-split}, where the green, blue and red blocks show the frontal slices used when computing the embeddings for the training, validation and testing data, respectively. The embeddings for the validation and testing data are $\Ye_\text{val} = \Ae_{::(S_\text{val}+1:S_\text{train}+S_\text{val})} \star \Xe_{::(S_\text{val}+1:S_\text{train}+S_\text{val})} \star \We$ and $\Ye_\text{test} = \Ae_{::(S_\text{val}+S_\text{test}+1:T)} \star \Xe_{::(S_\text{val}+S_\text{test}+1:T)} \star \We$, respectively.
For link prediction, we use the same embeddings, with the only difference that the embedding blocks contain $S_\text{train}-1$ slices. This is necessary since we want to use information up to time $t$ to predict edge existence at time $t+1$. 
Preliminary experiments with 2-layer architectures did not show convincing improvements in performance. We believe this is due to the fact that the datasets only have two features, and that a 1-layer architecture therefore is sufficient for extracting relevant information in the data.

For training, we use the cross entropy loss function:
\begin{equation} \label{eq:loss}
    \text{loss} = - \sum_{t} \sum_{(m,n) \in E_t} \sum_{c=1}^C \alpha_{c} f(m,n,t)_{c} \log(p(m,n,t)_{c}),
\end{equation}
where $\alpha \in \Rb^{C}$ is a vector summing to 1 which contains the weight of each class. For edge classification, $f(m,n,t) \in \Rb^{C}$ is a one-hot vector encoding the true class of the edge $(m,n)$ at time $t$. For link prediction, $f(m,n,t) \in \Rb^2$ is also a one-hot vector, but now encoding if the edge is an existing or non-existing edge.   
As appropriate, we weigh the minority class more heavily in the loss function for skewed datasets, and treat $\alpha$ as a hyperparameter. See Section~\ref{sec:hyperparameter-tuning} of the supplement for further details on the experiment setup, including the training setup and how hyperparameter tuning is done.

The experiments are implemented in PyTorch with some preprocessing done in Matlab. Our code is available at \url{https://github.com/IBM/TM-GCN}. In the experiments, we use an edge life of $l=10$, a bandwidth $b = 20$, and $F' = 6$ output features. For \ours, we consider two variants of the $\Mbf$ matrix (M1 and M2); see the supplement for details.

\begin{figure*}
	\includegraphics[width=.3\textwidth,trim={1.3cm 1.1cm 1.0cm 0.55cm}]{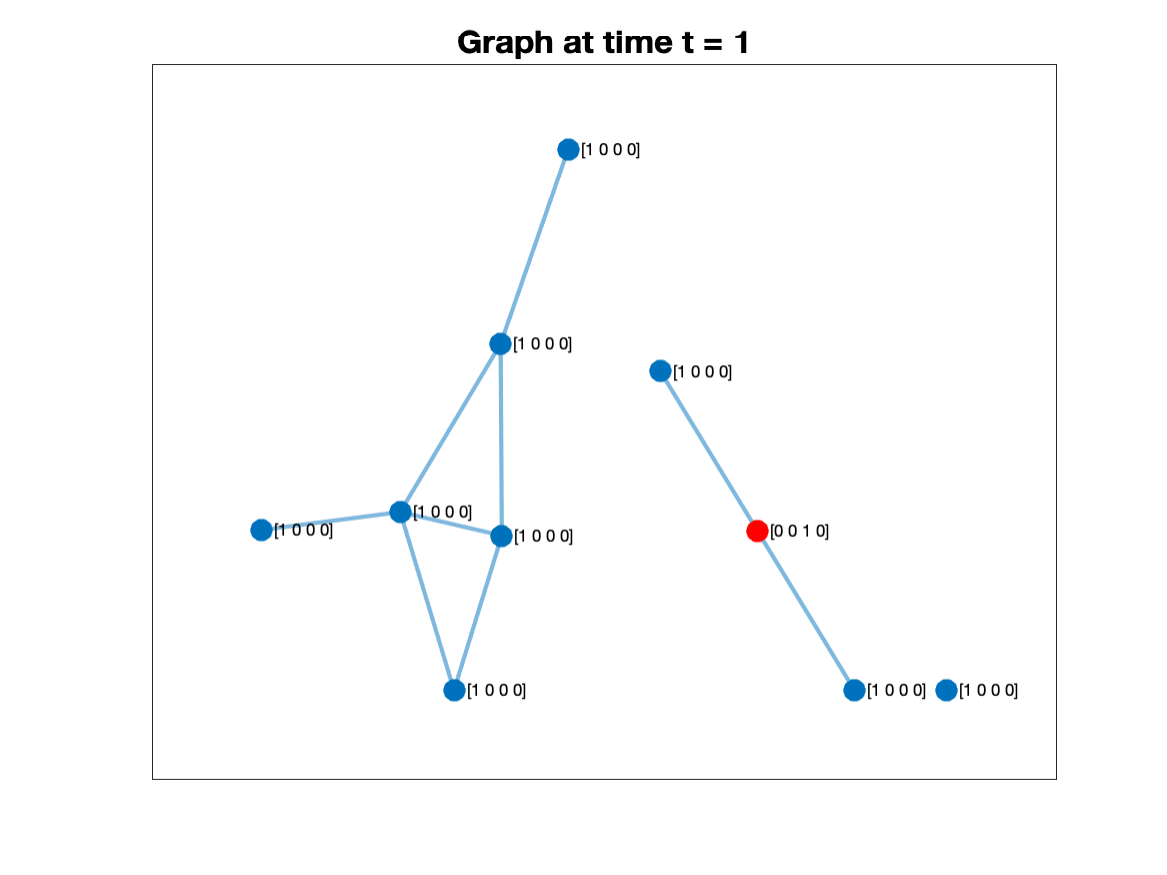}
	\includegraphics[width=.3\textwidth,trim={1.1cm 1.1cm 1.0cm 0.55cm}]{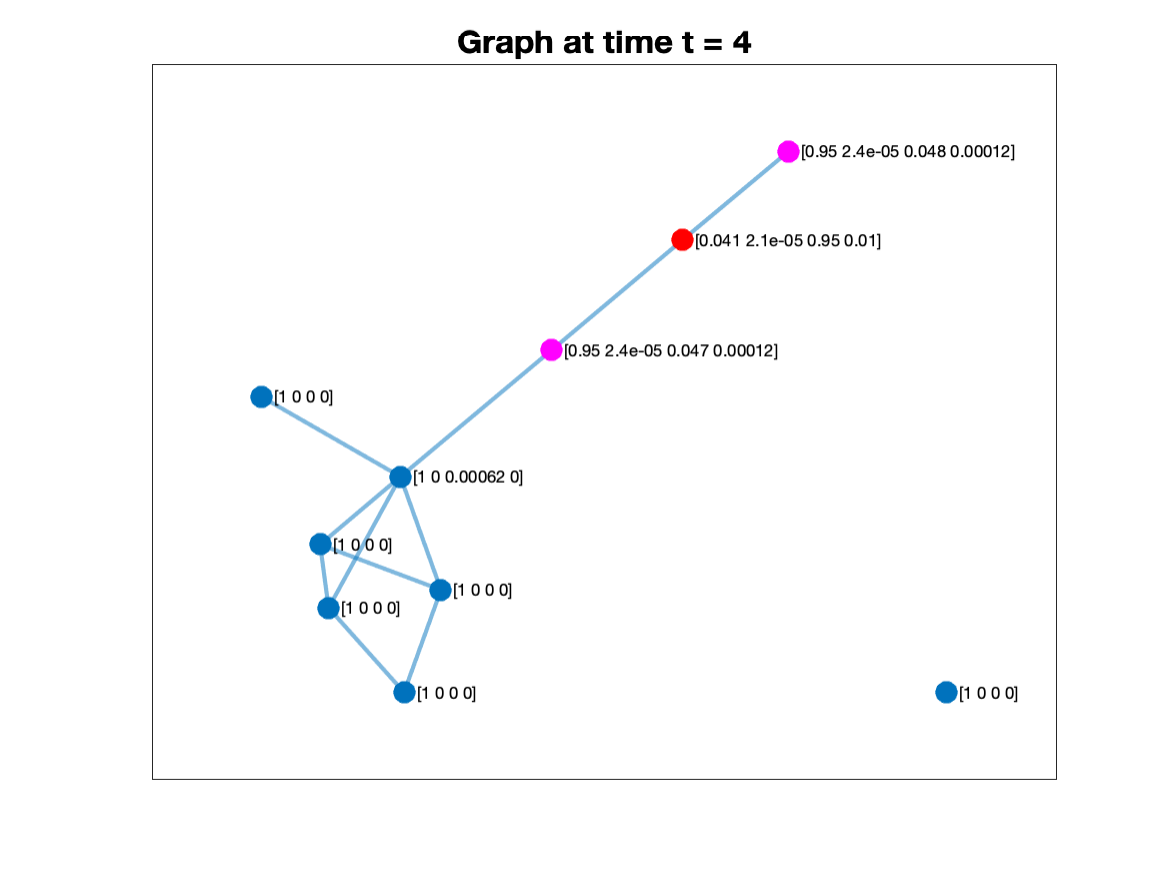}
	\includegraphics[width=.3\textwidth,trim={1.1cm 1.1cm 1.1cm 0.55cm}]{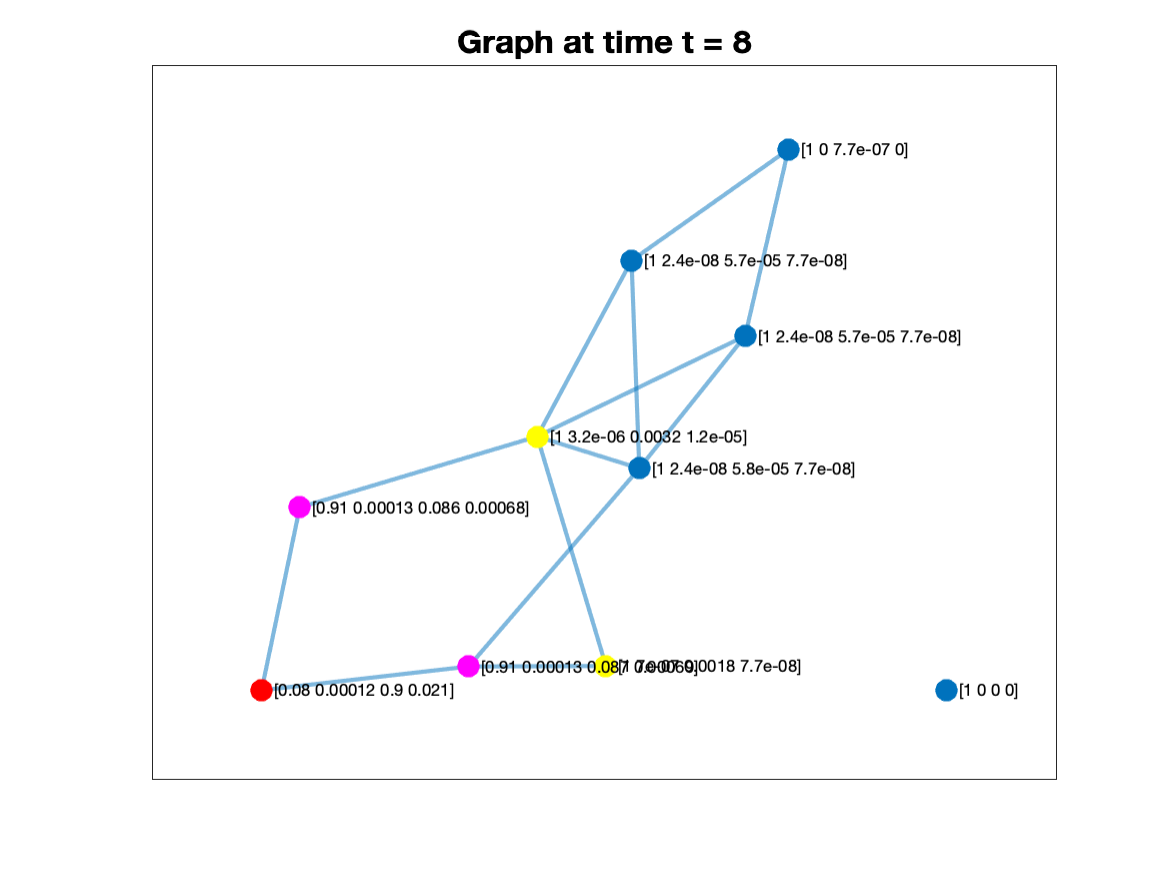}
	\caption{Graphical SEIR model disease transmission visualization.}
	\label{fig:Covid_graph}
\end{figure*}

We compare our method with three other methods. The first one is a variant of the WD-GCN by \cite{manessi2019}, which they specify in Equation~(8a) of their paper. For the LSTM layer in their description, we use $6$ output features instead of $N$. This is to avoid overfitting and make the method more comparable to ours which uses 6 output features. 
The second method is a 1-layer variant of EvolveGCN-H by \cite{pareja2019}. 
The third method is a simple baseline which uses a 1-layer version of the GCN by \cite{kipf2016}. It uses the same weight matrix $\Wbf$ for all temporal graphs. Both EvolveGCN-H and the baseline GCN use 6 output features as well.
We use the prediction model \eqref{eq:prediction-model} as the final layer in all models we compare.

Tables~\ref{tab:results-asymmetric} and \ref{tab:link-prediction} show the results for edge classification and link prediction, respectively. For edge classification, our method outperforms the other methods on the two bitcoin datasets and the chess dataset, with WD-GCN performing best on the Reddit dataset. For link prediction, our method outperforms the other methods on the SBM, bitcoin and chess datasets. For Reddit, our method performs worse than the other methods. 
Results from some additional experiments are provided in Section~\ref{sec:additional-experiments} of the supplement.

\subsection{COVID-19 Application}
One of the primary challenges related to the COVID-19 pandemic has been the issue of identifying early  the individuals who are infected (ideally before they display symptoms) and prescribe testing. Here, we demonstrate how we can potentially use GNNs on contact tracing data to  achieve this.

Contact tracing, a process where interactions between individuals (infected and others) are carefully tracked, has been shown the be an effective method for managing the spread of COVID-19. A variety of contact tracing methodologies have been used today around the world, see~\cite{alsdurf2020covi,ubaru2020dynamic} for lists. Recently, Ubaru et.~al~\cite{ubaru2020dynamic} presented a probabilistic graphical  SEIR  epidemiological model (Susceptible, Exposed, Infected, Recovered) to describe the dynamics of the disease transmission. Their model considers a dynamic graph (with individuals as nodes) that accounts for the interactions between individuals obtained from contact tracing, and uses a stochastic diffusion-reaction model to describe the disease transmission over the graph.  
\begin{table}
\centering
     \caption{COVID-19 Data: Mean absolute error and error ratio for infection state $I$ prediction.  \label{tab:results-covid}}
    \begin{tabular}{lcc}
    \hline
    Methods   & \multicolumn{2}{c}{COVID-19 Dataset}\\
    \cline{2-3}
                   & Error & Ratio         \\
    \hline
    WD-GCN                  & 1.667            & 0.337               \\
    EvolveGCN               & 4.969            & 0.912            \\
    \ours     & \textbf{1.466}   & \textbf{0.278}   \\
    \hline
    \end{tabular}

\end{table}

The novel SEIR model in~\cite{ubaru2020dynamic} considers the graph Laplacian $\Lbf_t$ (from contact tracing data) at each time $t$ and describes the evolution of the state $\{S,E,I,R\}$ for each node/individual.
Figure~\ref{fig:Covid_graph} illustrates the state $\{S,E,I,R\}$ evolution as defined by the model on a sample dynamic graph with 10 individuals (for easy visualization). We see how the infection (one individual as red node in first graph) transmits, we have magenta nodes with $I >0.04$, and the yellow  nodes with $I >0.002$, and we note the interactions and the state change over time.   Here, we show how we can use dynamic GNNs to predict the infection state $I$ at time $T+1$, using only the dynamic graphs up to time $T$, when the true SEIR model is unknown. 

We consider a simulation with $N=1000$ individuals and total time $T=100$. We simulate the contact tracing dynamic graph as in~\cite{ubaru2020dynamic}, and assume at each time $t$ a small number of individuals are tested at random for both  IgM (if positive state $I$ is set to 1) and IgG (state $R$ is set to 1) antigen tests. The state of the remaining  individuals are determined by the SEIR model. We train the dynamic GNNs on the first $T=80$ time instances and test the GNNs on the remaining $20$ time instances. Our goal is to train a GNN that learns the   relation between the contact tracing graphs and the infection state (some exact values for those who were tested and others from the SEIR model), in order to \emph{better  predict} the  individuals' state $I$ at time $t+1$ than just using the SEIR model. Table~\ref{tab:results-covid} gives the mean absolute error and error ratio obtained by the three dynamic GNNs for infection state $I$ prediction on the test time instances. We note that, the proposed \ours~ yields best results among the three methods, since it has a better time awareness (explicitly considers $b$ previous time instances via the M-product) than  others. Using such predictions,  we can issue early warnings to individuals who are infected and prescribe testing.

\section{Conclusion}

We have presented a novel approach for dynamic graph embedding which leverages the tensor M-product framework. 
We used it for edge classification and link prediction in experiments on five datasets, where it performed competitively compared to state-of-the-art methods. We also demonstrated the method's effectiveness in an important application related to the COVID-19 pandemic.
Future research directions include further developing the theoretical guarantees for the method, investigating optimal structure and learning of the transform matrix $\Mbf$, using the method for other prediction tasks, and investigating how to utilize deeper architectures for dynamic graph learning.

\section*{Acknowledgments}

We thank Stephen Becker and Lingfei Wu for helpful discussions and feedback.
We also thank the reviewers for their insightful comments and suggestions.
Kilmer was partially supported by a grant from IBM T.J.\ Watson and by Tufts T-Tripods Institute under NSF HDR grant CCF-1934553.
Avron was supported by Israel Science Foundation grant 1272/17 and by an IBM Faculty Award.

{
\bibliography{zotero-library}
\bibliographystyle{abbrv}
}
\newpage
\clearpage
\appendix

\onecolumn
\begin{center}
   \bf \large{Supplementary Material: \\Dynamic Graph Convolutional Networks Using the Tensor M-Product}
\end{center}
\vskip 0.3in

\section{Links to Datasets} \label{sec:datasets}

\begin{itemize}
    \item The Bitcoin Alpha dataset is available at  \url{https://snap.stanford.edu/data/soc-sign-bitcoin-alpha.html}.
    \item The Bitcoin OTC dataset is available at  \url{https://snap.stanford.edu/data/soc-sign-bitcoin-otc.html}.
    \item The Reddit dataset is available at  \url{https://snap.stanford.edu/data/soc-RedditHyperlinks.html}. Note that we use the dataset with hyperlinks in the body of the posts.
    \item The chess dataset was originally downloaded from \url{http://konect.uni-koblenz.de/networks/chess}. 
    This link no longer works, and the Koblenz Network Collection appears to no longer be available online. 
    We therefore added the chess dataset to the TM-GCN GitHub repository at \url{https://github.com/IBM/TM-GCN}.
    \item SBM (structured block matrix) was generated using the code \url{https://github.com/palash1992/DynamicGEM/}. We used 1000 nodes, 8 communities (blocks) and 50 time instances.
\end{itemize}

\section{Further Details on the Experiment Setup} \label{sec:hyperparameter-tuning}

When partitioning the data into $T$ graphs, as described in Section~\ref{sec:experiments}, if there are multiple data points corresponding to an edge $(m,n)$ for a given time step $t$, we only add that edge once to the corresponding graph and set the label equal to the sum of the labels of the different data points. E.g., if bitcoin user $m$ makes three transactions to $n$ during time step $t$ with ratings $10$, $2$, $-1$, then we add a single edge $(m,n)$ to graph $t$ with label $10+2-1 = 11$.

\subsection{Edge Classification}

For training, we run gradient descent with a learning rate of 0.01 and momentum of 0.9 for 10,000 iterations. For each 100 iterations, we compute and store the performance of the model on the validation data.
The weight vector $\alpha$ in the loss function \eqref{eq:loss} is treated as a hyperparameter in the bitcoin and Reddit experiments. Since these datasets all have two edge classes, let $\alpha_0$ and $\alpha_1$ be the weights of the minority (negative) and majority (positive) classes, respectively. Since these parameters add to 1, we have $\alpha_1 = 1-\alpha_0$. For all methods, we repeat the bitcoin and Reddit experiments once for each $\alpha_0 \in \{0.75, 0.76, \ldots, 0.95\}$. For each model and dataset, we then find the best stored performance of the model on the validation data across all $\alpha_0$ values. We then treat the corresponding model as the trained model, and report its performance on the test data. 
The results for the chess experiment are computed in the same way, but only for a single vector $\alpha = [1/3, \; 1/3, \; 1/3]$.

\subsection{Link Prediction}

For training, we run gradient descent with a learning rate of 0.01 and momentum of 0.9 for 1,000 iterations. For each 100 iterations, we compute and store the performance of the model on the validation data. We used $\alpha_0 = 0.90$ in our experiments, where $\alpha_0$ is the weight of the class corresponding to existing edges. 

\subsection{Definition of Performance Measures}

Suppose we are classifying $N$ objects into $C$ classes. Let $\ubf \in \{1,2,\ldots,C\}^{N}$ be a vector containing our computed classification, and let $\vbf \in \{1,2,\ldots,C\}^N$ be a vector which contains the true classes. Furthermore, let $k$ denote the class we are interested in identifying (i.e., negative edges in bitcoin and Reddit edge classification problems). Then, the F1 score is defined as follows: 
\begin{equation}
    \text{F1 score} = 2 \cdot \frac{\text{precision} \cdot \text{recall}}{\text{precision} + \text{recall}},
\end{equation}
where
\begin{equation}
\begin{aligned}
    \text{precision} &\defeq \frac{\text{true positive}}{\text{true positive} + \text{false positive}}, \\
    \text{recall} &\defeq \frac{\text{true positive}}{\text{true positive} + \text{false negative}}, \\
    \text{true positive} &\defeq |\{n \in \{1, 2, \ldots, N\} : \ubf_n = \vbf_n = k\}|, \\
    \text{false positive} &\defeq |\{n \in \{1, 2, \ldots, N\} : \ubf_n = k, \vbf_n \neq k\}|, \\
    \text{false negative} &\defeq |\{n \in \{1, 2, \ldots, N\} : \ubf_n \neq k, \vbf_n = k\}|. \\
\end{aligned}
\end{equation}

For the accuracy in the edge classification experiment on the chess dataset, we simply compute it as the proportion of correctly labeled edges.

For computing mean average precision, we use the \verb|average_precision_score| function in the Scikit-learn library. As input, we use a vector containing the probabilities for the class of interest (i.e., the class corresponding to existing edges in link prediction) generated by the output model \eqref{eq:prediction-model}.

\subsection{Choice of M Matrix} We consider two variants of the lower banded triangular $\Mbf$ matrices in our experiments. 
Specifically, in the first matrix $M1$, the entries of $\Mbf$ are set to
\begin{equation} \label{eq:def-M1}
    \Mbf_{tk} \defeq 
    \begin{cases}
        \frac{1}{\min(b, t)}    & \text{if } \max(1, t-b+1) \leq k \leq t,\\
        0                       & \text{otherwise},
    \end{cases}
\end{equation}
which implies that $\sum_{k} \Mbf_{tk} = 1$ for each $t$. However, this transform gives equal weights to all the previous $b$ time instances.
In the second matrix (M2), we have the entries as:
\begin{equation} \label{eq:def-M2}
    \Mbf_{tk} \defeq 
    \begin{cases}
        \frac{1}{k}    & \text{if } \max(1, t-b+1) \leq k \leq t,\\
        0                       & \text{otherwise}.
    \end{cases}
\end{equation}
This transform gives exponential weights to the $b$ previous time instances, weighing recent ones more, and the weights decay exponentially for older time instances.
\section{Additional Experimental Results} \label{sec:additional-experiments}

Since the graphs in the considered datasets are directed, we also investigate the impact of symmetrizing the adjacency matrices, where the symmetrized version of an adjacency matrix $\Abf$ is defined as $\Abf_{\text{sym}} \defeq 1/2(\Abf + \Abf^\top)$. Table~\ref{tab:results-symmetric} shows the results.
Our method outperforms the other methods on the Bitcoin OTC dataset and the chess dataset, and performs similarly but slightly worse than the best performing methods on the Bitcoin Alpha and Reddit datasets. Overall, it seems like symmetrizing the adjacency matrices leads to lower performance.

\begin{table}[h!]
\centering
    \caption{Results for edge classification when adjacency matrices have been symmetrized. Performance measures are F1 score\tdag or accuracy\tast. A higher value is better.\label{tab:results-symmetric}}
    \begin{tabularx}{0.8\textwidth}{lYYYY}
    \toprule
    & \multicolumn{4}{c}{Dataset}\\
    \cmidrule(l){2-5}
    Method                  & Bitcoin OTC\tdag  & Bitcoin Alpha\tdag& Reddit\tdag       & Chess\tast        \\
    \midrule
    WD-GCN                  & 0.1009            & 0.1319            & \textbf{0.2173}   & 0.4321            \\
    EvolveGCN               & 0.0913            & \textbf{0.2273}   & 0.1942            & 0.4091            \\
    GCN                     & 0.0769            & 0.1538            & 0.1966            & 0.4369            \\
    TM-GCN - M1    			& \textbf{0.3103}   & 0.2207            & 0.2071            & \textbf{0.4713}   \\
    \bottomrule
    \end{tabularx}
\end{table}

\section{Additional Details and  Proofs} \label{sec:proofs}
Here, we give additional details regarding the tensor M-product framework. We also present a few additional theoretical results and proofs.

\subsection{Additional Details}

A benefit of the tensor M-product framework is that many standard matrix concepts can be generalized in a straightforward manner. Definitions~\ref{def:f-diagonal}--\ref{def:orthogonal-tensor} extend the matrix concepts of diagonality, identity, transpose and orthogonality to tensors \cite{braman2010, kilmer2013}.
\begin{Definition}[f-diagonal] \label{def:f-diagonal}
    A tensor $\Xe \in \Rb^{N \times N \times T}$ is said to be \emph{f-diagonal} if each frontal slice $\Xe_{::t}$ is diagonal.
\end{Definition}
\begin{Definition}[Identity tensor] \label{def:identity-tensor}
    Let $\hat{\Ie} \in \Rb^{N \times N \times T}$ be defined facewise as $\hat{\Ie}_{::t} = \Ibf$, where $\Ibf$ is the matrix identity. The M-product \emph{identity tensor} $\Ie \in \Rb^{N \times N \times T}$ is then defined as $\Ie \defeq \hat{\Ie} \times_3 \Mbf^{-1}$. 
\end{Definition}
\begin{Definition}[Tensor transpose]  \label{def:tensor-inverse}
    The transpose of a tensor $\Xe$ is defined as $\Xe^\top \defeq \Ye \times_3 \Mbf^{-1}$, where $\Ye_{::t} = (\Xe \times_3 \Mbf)_{::t}^\top$ for each $t \in \{1,\ldots,T\}$.
\end{Definition}

\begin{Definition}[Orthogonal tensor] \label{def:orthogonal-tensor}
    A tensor $\Xe \in \Rb^{N \times N \times T}$ is said to be \emph{orthogonal} if $\Xe \star \Xe^\top = \Xe^\top \star \Xe = \Ie$.
\end{Definition}

Leveraging these concepts, a tensor eigendecomposition can now be defined \cite{braman2010, kilmer2013}:
\begin{Definition}[Tensor eigendecomposition] \label{def:tensor-eigendecomposition}
    Let $\Xe \in \Rb^{N \times N \times T}$ be a tensor and assume that each frontal slice $(\Xe \times_3 \Mbf)_{::t}$ is symmetric. We can then eigendecompose these as $(\Xe \times_3 \Mbf)_{::t} = \hat{\Qe}_{::t} \hat{\De}_{::t} \hat{\Qe}_{::t}^{\top}$, where  $\hat{\Qe}_{::t} \in \Rb^{N \times N}$ is orthogonal and $\hat{\De}_{::t} \in \Rb^{N \times N}$ is diagonal. We assume that the eigenvalues along the diagonal of each $\hat{\De}_{::t}$ are ordered in descending order, i.e., $\hat{\De}_{nnt} \geq \hat{\De}_{mmt}$ whenever $n < m$. The \emph{tensor eigendecomposition} of $\Xe$ is defined as
    $
        \Xe \defeq \Qe \star \De \star \Qe^\top,
    $
    where $\Qe \defeq \hat{\Qe} \times_3 \Mbf^{-1}$ is orthogonal, and $\De \defeq \hat{\De} \times_3 \Mbf^{-1}$ if f-diagonal.
\end{Definition}

\begin{figure}[htb!]
  \centering
  \includegraphics[width=.3\columnwidth]{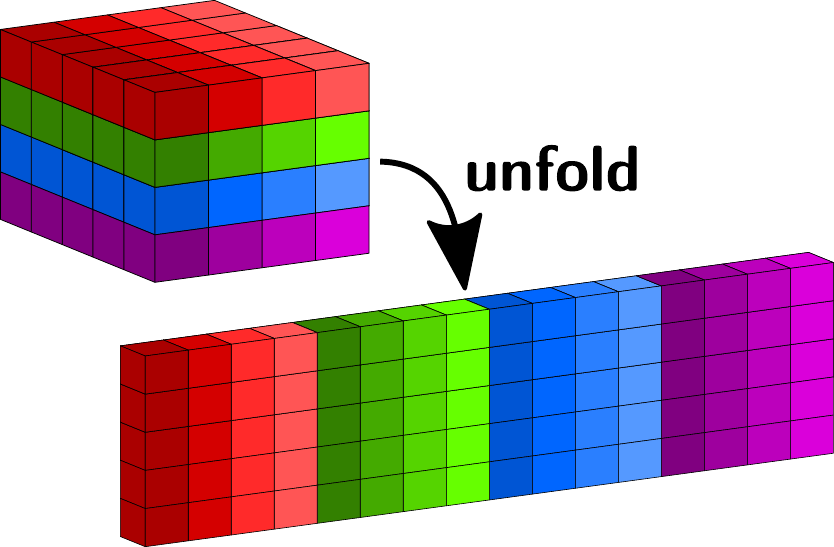}\quad \quad
  \includegraphics[width=.5\linewidth]{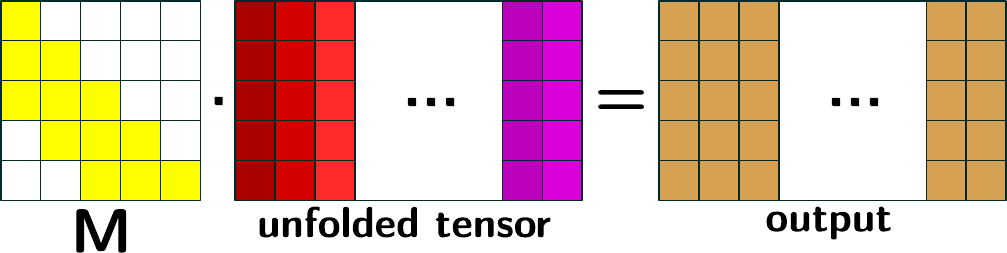}
  \caption{Illustration of (left) unfold operation applied to $4 \times 4 \times 5$ tensor, and (right) matrix product between $\Mbf$ and the unfolded tensor.}
  \label{fig:unfold}
\end{figure}

\paragraph{Illustration:} Figure~\ref{fig:unfold} (left) illustrates the unfolding operation. Figure~\ref{fig:unfold} (right) shows how, once unfolded, the matrix $\unfold(\Xe)$ is multiplied from the left by $\Mbf$, which has a lower triangular banded structure. The output is then folded back up into a tensor by doing the inverse operation of that illustrated in Figure~\ref{fig:unfold}.

\subsection{Additional Results}
Here, we present a few more results related to our analysis in Section~\ref{sec:theoretical-analysis}. 
Much like the spectrum of a normalized graph Laplacian is contained in $[0,2]$, the tensor spectrum of $\Le$ satisfies a similar property when $\Mbf$ is chosen appropriately.
\begin{proposition}[Spectral bound] \label{prop:Ae-spectrum}
    The entries of $\hat{\De} = \De \times_3 \Mbf$ lie in $[0,2]$ for the first $\Mbf$ matrix $M1$.
\end{proposition}

\begin{proof}
	Each $\Ae_{::t}$ has a spectrum contained in $[-1,1]$. Since $\Ae_{::t}$ is symmetric, it follows that $\|\Ae_{::t}\|_2 \leq 1$. Consequently, 
	\begin{equation}
	\|(\Ae \times_3 \Mbf)_{::t}\|_2 = \Big\| \sum_{j=1}^T \Mbf_{tj} \Ae_{::j} \Big\|_2 \leq \sum_{j=1}^T |\Mbf_{tj}| \|\Ae_{::j}\|_2 \leq 1,
	\end{equation}
	where we used the fact that $\sum_j |\Mbf_{tj}| = 1$. So since the frontal slices $(\Ae \times_3 \Mbf)_{::t}$ are symmetric, they each have a spectrum in $[-1,1]$. It follows that each frontal slice
	\begin{equation}
	(\Le \times_3 \Mbf)_{::t} = \Ibf - (\Ae \times_3 \Mbf)_{::t}
	\end{equation}
	has a spectrum contained in $[0,2]$, which means that the entries of $\hat{\De}$ all lie in $[0,2]$.
\end{proof}

Following the work by \cite{kilmer2013}, three-dimensional tensors in $\Rb^{M \times N \times T}$ can be viewed as operators on $N \times T$ matrices, with those matrices ``twisted'' into tensors in  $\Rb^{N \times 1 \times T}$. With this in mind, we define a tensor variant of the graph Fourier transform.
\begin{Definition}[Tensor-tube M-product]
    Let $\Xe \in \Rb^{I \times J \times T}$ and $\thetabf \in \Rb^{1 \times 1 \times T}$. Analogously to the definition of the matrix-scalar product, we define $\Xe \star \thetabf$ via $(\Xe \star \thetabf)_{ij:} \defeq \Xe_{ij:} \star \thetabf$.
\end{Definition}
\begin{Definition}[Tensor graph Fourier transform]
    Let $\Xe \in \Rb^{N \times F \times T}$ be a tensor, and let $\Qe$ be defined as in Definition~\ref{prop:existence-eigendecomp}. We define a \emph{tensor graph Fourier transform} $F$ via $F(\Xe) \defeq \Qe^\top \star \Xe \in \Rb^{N \times F \times T}$.
\end{Definition}
This is analogous to the definition of the matrix graph Fourier transform. This defines a convolution like operation for tensors similar to spectral graph convolution~\cite{bruna2013}.
Each lateral slice $\Xe_{:j:}$ is expressible in terms of the set $\{\Qe_{:n:}\}_{n=1}^N$ as follows:
\begin{equation}
    \Xe_{:j:} = \Qe \star \Qe^{\top} \star \Xe_{:j:} = \sum_{n=1}^N \Qe_{:n:} \star (\Qe^\top \star \Xe_{:j:})_{n1:},
\end{equation}
where each $(\Qe^\top \star \Xe_{:j:})_{n1:} \in \Rb^{1 \times 1 \times T}$ can be considered a tubal scalar. In fact, the lateral slices $\Qe_{:n:}$ form a basis for the set $\Rb^{N \times 1 \times T}$ with product $\star$.

In the following, $\|\cdot\|$ will denote the Frobenius norm (i.e., the square root of the sum of the elements squared) of a matrix or tensor, and $\|\cdot\|_2$ will denote the matrix spectral norm.
We first provide a few further results that clarify the algebraic properties of the M-product. 
Let $\Rb^{1 \times 1 \times T}$ denote the set of $1 \times 1 \times T$ tensors. 
Similarly, let $\Rb^{N \times 1 \times T}$ denote the set of $N \times 1 \times T$ tensors. 
Under the M-product framework, the set $\Rb^{1 \times 1 \times T}$ plays a role similar to that played by scalars in matrix algebra. 
With this in mind, the set $\Rb^{N \times 1 \times T}$ can be seen as the length-$N$ vectors consisting of tubal elements of length $T$.
Propositions~\ref{prop:ring} and \ref{prop:free-module} make this more precise.
\begin{proposition}[Proposition 4.2 in \cite{kernfeld2015}] \label{prop:ring}
    The set $\Rb^{1 \times 1 \times T}$ with product $\star$, which is denoted by $(\star, \Rb^{1 \times 1 \times T})$, is a commutative ring with identity.
\end{proposition}

\begin{proposition}[Theorem 4.1 in \cite{kernfeld2015}] \label{prop:free-module}
    The set $\Rb^{N \times 1 \times T}$ with product $\star$, which is denoted by $(\star, \Rb^{N \times 1 \times T})$, is a free module over the ring $(\star, \Rb^{1 \times 1 \times T})$.
\end{proposition}
A free module is similar to a vector space. Like a vector space, it has a basis. Proposition~\ref{prop:module-basis} shows that the lateral slices of $\Qe$ in the tensor eigendecomposition form a basis for $(\star, \Rb^{N \times 1 \times T})$, similarly to how the eigenvectors in a matrix eigendecomposition form a basis.
\begin{proposition} \label{prop:module-basis}
    The lateral slices $\Qe_{:n:} \in \Rb^{N \times 1 \times T}$ of $\Qe$ in Definition~\ref{def:tensor-eigendecomposition} form a basis for $(\star, \Rb^{N \times 1 \times T})$.
\end{proposition}
\begin{proof}
    Let $\Xe \in \Rb^{N \times 1 \times T}$. Note that
    \begin{equation}
        \Xe = \Ie \star \Xe = \Qe \star \Qe^\top \star \Xe = \sum_{n=1}^N \Qe_{:n:} \star \Ve_{n1:},
    \end{equation}
    where $\Ve \defeq \Qe^\top \star \Xe \in \Rb^{N \times 1 \times T}$. So the lateral slices of $\Qe$ are a generating set for $(\star, \Rb^{N \times 1 \times T})$. Now suppose
    \begin{equation}
        \sum_{n=1}^N \Qe_{:n:} \star \Se_{n1:} = \zerobf,
    \end{equation}
    for some  $\Se \in \Rb^{N \times 1 \times T}$. Then $\zerobf = \Qe \star \Se$, and consequently
    \begin{equation}
         \zerobf = (\Qe \times_3 \Mbf) \triangleop (\Se \times_3 \Mbf).
    \end{equation}
    Since each frontal face of $\Qe \times_3 \Mbf$ is an invertible matrix, this implies that each frontal face of $\Se \times_3 \Mbf$ is zero, and hence $\Se = \zerobf$. So the lateral slices of $\Qe$ are also linearly independent in $(\star, \Rb^{N \times 1 \times T})$.
\end{proof}

\subsection{Proofs of Propositions in the Main Text}

\begin{proof}\textit{(Proposition~\ref{prop:existence-eigendecomp})}
    Since each adjacency matrix $\Abf^{(t)}$ and each $\Ie_{::t}$ is symmetric, each frontal slice $\Le_{::t}$ is also symmetric. Consequently,
    \begin{equation}
        (\Le \times_3 \Mbf)_{ij:} = \Le_{ij:} \times_3 \Mbf = \Le_{ji:} \times_3 \Mbf = (\Le \times_3 \Mbf)_{ji:},
    \end{equation}
    so each frontal slice of $\Le \times_3 \Mbf$ is symmetric, and therefore $\Le$ has an eigendecomposition.
\end{proof}

\begin{lemma} \label{lemma:inequality}
    Let $\Xe \in \Rb^{M \times N \times T}$ and let $\Mbf \in \Rb^{T \times T}$ be invertible. Then 
    \begin{equation}
        \|\Xe\| \leq \|\Mbf^{-1}\|_2 \|\Xe \times_3 \Mbf\|.
    \end{equation}
\end{lemma}
\begin{proof}
    We have
    \begin{equation}
    \begin{aligned}
        &\|\Xe\| = \|(\Xe \times_3 \Mbf) \times_3 \Mbf^{-1}\| = \|\Mbf^{-1} \unfold(\Xe \times_3 \Mbf)\| \\
        &\leq \|\Mbf^{-1}\|_2 \|\unfold(\Xe \times_3 \Mbf)\| = \|\Mbf^{-1}\|_2 \| \Xe \times_3 \Mbf \|,
    \end{aligned}
    \end{equation}
    where the inequality is a well-known relation that holds for all matrices.
\end{proof}

\begin{proof}\textit{(Proposition~\ref{prop:tensor-polynomial-approx})}
	We show the proof for the case when $\Mbf$ is defined as in \eqref{eq:def-M1}.
	However, this proof can easily be adapted to when $\Mbf$ is defined as in \eqref{eq:def-M2} by adapting the interval $[0,2]$ in \eqref{eq:sup} appropriately.
	
    By Weierstrass approximation theorem, there exists an integer $K$ and a set $\{\hat{\thetabf}^{(k)}\}_{k=1}^K \subset \Rb^{1 \times 1 \times T}$ such that for all $t \in \{1,2,\ldots,T\}$,
    \begin{equation} \label{eq:sup}
        \sup_{x \in [0,2]} \Big| f^{(t)}(x) - \sum_{k=0}^K x^k \hat{\thetabf}^{(k)}_{11t} \Big| < \frac{\varepsilon}{\|\Mbf^{-1}\|_2 \sqrt{NT}}.
    \end{equation}
    Let $\thetabf^{(k)} \defeq \hat{\thetabf}^{(k)} \times_3 \Mbf^{-1}$.
    Note that if $m \neq n$, then
    \begin{equation}
        \Big(\sum_{k=0}^K \De^{\star k} \star \thetabf^{(k)}\Big)_{mn:} = \sum_{k=0}^K ((\hat{\De}^{\triangleop k})_{mn:} \times_{3} \Mbf^{-1}) \star \thetabf^{(k)} = \zerobf = g(\De)_{mn:},
    \end{equation}
    since $\hat{\De} = \De \times_3 \Mbf$ is f-diagonal.
    So
    \begin{equation}
    \begin{aligned}
        &\Big\| g(\De) - \sum_{k=0}^K \De^{\star k} \star \thetabf^{(k)}\Big\|^2 
        = \sum_{n=1}^N \Big\| g(\De)_{nn:} - \sum_{k=0}^K (\De^{\star k})_{nn:} \star \thetabf^{(k)}\Big\|^2 \\
        & \leq \| \Mbf^{-1} \|^2_2 \sum_{n=1}^N \Big\| g(\De)_{nn:} \times_3 \Mbf - \sum_{k=0}^K ((\De \times_3 \Mbf)^{\triangleop k})_{nn:} \triangleop \hat{\thetabf}^{(k)} \Big\|^2 \\
        & = \| \Mbf^{-1} \|^2_2 \sum_{n=1}^N \sum_{t=1}^T \Big| f^{(t)}((\De \times_3 \Mbf)_{nnt}) - \sum_{k=0}^K (\De \times_3 \Mbf)^{k}_{nnt} \hat{\thetabf}^{(k)}_{11t} \Big|^2 \\
        & < \varepsilon^2,
    \end{aligned}
    \end{equation}
    where the first inequality follows from Lemma~\ref{lemma:inequality}, and the last inequality follows since $(\De \times_3 \Mbf)_{nnt} \in [0,2]$ due to Proposition~\ref{prop:Ae-spectrum} (that proposition can easily be adapted to the case when $\Mbf$ is defined as in \eqref{eq:def-M2}).
    Taking square roots completes the proof.
\end{proof}
 Note that, the above proof holds even when we do not apply  the inverse transform $\Mbf^{-1}$, since $\Mbf^{-1}$ only shows up as the norm $\| \Mbf^{-1} \|_2$ at the end and the whole proof is still consistent  without it.
\end{document}